# Context-Dependent Similarity

Yizong Cheng

Department of Computer Science
University of Cincinnati
Cincinnati, OH 45221-0008
yizong.cheng@uc.edu

## Abstract

Numerical similarity measures are used to describe the relative ranks of the similarity of objects or cases in many artificial intelligent systems. These measures are usually absolute and context-independent. On the other hand, humans perceive context-dependent similarity. That is, the ranking of similarity between pairs of objects is varying under a changing context. We consider this problem as the construction of numerical formulas that satisfy certain axioms and criteria. An entropy-related formula is proposed and its implementation in a changing environment is considered. A demonstration of this formula on a well-known context-dependent similarity assessment is given.

## 1 Introduction

Many artificial intelligence models depend on the similarity measure on objects or cases. For example, concept formation, similarity-based learning, and analogical reasoning use similarity to classify, store and retrieve data. The similarity between two cases or two objects is often computed based on the descriptions of the cases or objects, which is often represented as a set of attribute values. For example, the Hamming distance is the number of attributes that have different values in two cases. In most existing works, the similarity measures solely depend on the attribute values of the cases, and is invariant with respect to the context changes. This, of course, has become the target of some criticisms from researchers in experimental psychology.

The basic observations from psychologists are that if similarity represents a ranking on pairs of cases, then the ranking is varying under changing context and the history of observation. So the common practice that one imposes a fixed, global ranking to the similarity is an example of oversimplification.

In this paper, we study some of the major criticisms and propose our dynamic model for adjusting attribute weights and thus the similarity computation formulas. Both an axiomatical and an entropy-oriented approaches are used to derive satisfactory formulas. Demonstration and implementation are also considered.

## 2 Similarity Is Context-Dependent

The best-known demonstration of the unstability of similarity rating is from Tversky and Gati (1978). The subjects were asked to group *Austria* with one of three other countries based on their similarity. When the other three countries were *Sweden, Poland*, and *Hungary*, more subjects grouped *Austria* with *Sweden* then with *Hungary*. However, when the three countries were *Sweden, Norway*, and *Hungary*, more subjects grouped *Austria* with *Hungary*. The explanation of the phenomenon is that the context determines and changes the way similarity is perceived.

The similarity between two cases can either be perceived holistically, or computed as a function of the differences of elementary attributes. Treisman and Gelade's (1980) experiment and almost all the models support the latter. On the other hand, researchers agree that the selection of elementary attributes is never permanent in the process of development, and the function relating them with similarity is dynamic accordingly (Vosniadou and Ortony, 1989). Attributes can be perceptual, cognitively primitive, or conceptual, based on an underlying model that combines elementary attributes. For composite cases (e.g. medical cases), attributes can be temporal, structural, or relational. In different context, different attributes of a case or object draw different attention and thus should be weighted dif-ferently. One example was given in Barclay et al. (1974) where the weight and the sound of a piano become the most relevant attribute when the word *piano* occurs in the sentences "The man lifted the piano" and "The man tuned the piano", respectively.

It is important to study context-dependent models and computational formulas for similarity assessment. We will discuss the requirements for a such model and also propose a concrete one. In Section 5, the model will be used to explain the apparent inconsistency in Tversky and Gati's experiment.

## 3 Axioms and Formulas for Similarity Computation

In this section, we propose several basic axioms as properties that a computational model for the similarity measure should have, and then choose a formula that realizes these properties. To start from simple problems, we assume that attributes are boolean, and cases are represented by specifying values for all the attributes. The context is no more than the collection of cases we have. Formally, we assume that there are $n$ attributes and $m$ cases. The value of the $j$th attribute of the $i$th case is $c_{ij}$, and the $i$th case is represented as a vector of the value $c_{ij}$'s, $\mathbf{c}_i = [c_{i1}, \ldots, c_{in}]^T$. Let us assume that $c_{ij}$ is either 0 or 1.

The similarity between cases is computed from the similarity between the attribute values. Let $e(c_{ij}, c_{kj})$ be the *dissimilarity* between the values of the $j$th attribute for the $i$th and the $k$th cases. Since we have assumed that the attributes are boolean, $e(c_{ij}, c_{kj})$ can assume only four values. If we assume that similarity is symmetric (which is not assumed in many models, for example, the "contrast rule" of Tversky), that is, $e(a,b) = e(b,a)$, then two of the four values are the same. If we assume that the two boolean values of an attribute are equally important (which is the case in many models, for example, the Hopfield associative memory, but does not seem to be so in the original Hebbian learning), then the other two values are the same. Let us assume that $e(a,b) = 0$ if $a = b$ and $e(a,b) = 1$ if $a \neq b$. The Hamming distance between $\mathbf{c}_i$ and $\mathbf{c}_k$ is

$$d(\mathbf{c}_i, \mathbf{c}_k) = \sum_{j=1}^{n} e(c_{ij}, c_{kj}). \tag{1}$$

The statistics of the attribute values over all cases can be used as the representation of the context. Let the probability that the $j$th attribute assumes 1 be $p_j$. If the attributes are statistically independent, then the probability that both the $j$th and the $l$th attributes are 1 is $p_j p_l$. Now the first question is: If $\mathbf{c}_1$ differs from $\mathbf{c}_2$ *only* at the $j$th attribute, $\mathbf{c}_1$ differs from $\mathbf{c}_3$ *only* at the $l$th attribute, and we know $p_j$ and $p_l$, which one of $\mathbf{c}_2$ and $\mathbf{c}_3$ is more similar to $\mathbf{c}_1$? We can come up with a expression to clarify this question. The expression gives the distance between two cases when the only difference between them is the $j$th attribute. If this distance depends on $p_j$ only, then it is in the form of $h(p_j)$.

Because the choice of the probability of being 1 instead of 0 for the definition of $p_j$ is arbitrary, we want

$$h(a) = h(1-a) \tag{2}$$

where $a \in [0,1]$. This is the property of symmetry. Now the interval $[0,1]$ is folded into $[0,1/2]$, for $h$ is symmetric with respect to $1/2$. We only have to consider the values of $h$ on $[0,1/2]$. Now we can rephrase our question more specifically: Do you consider $\mathbf{c}_1$ to be closer to $\mathbf{c}_2$ than to $\mathbf{c}_3$, if $\mathbf{c}_1$ differs from $\mathbf{c}_2$ only at the $j$th attribute and $p_j$ is very close to 0.5, while $\mathbf{c}_1$ differs from $\mathbf{c}_3$ only at the $l$th attribute and $p_l$ is very close to 0 or 1? This problem must be solved by experimental psychologists. But, as to be seen in the next section, in order to explain the existing experimental results, we have to take the side that $\mathbf{c}_1$ is closer to $\mathbf{c}_3$. In other words, the $l$th attribute provides weaker support in similarity computation.

Other reasonable assumptions can be made, too. We may ask for a smooth $h$, that is, the derivative of $h$ is positive on $[0,1/2]$. Because of symmetry and smoothness, the derivative at $1/2$ should be 0. In this way, we have established a one-to-one correspondence between the $h$ values and the interval $[0,1/2]$, the range of $p_j$'s. If the distance between two cases, $d(\mathbf{c}_i, \mathbf{c}_k)$, is a function of $p_j$'s, it can also be represented as a function of $h(p_j)$'s. Because the order of attributes is arbitrary, The arguments of the function must be exchangeable. Finally, it should be monotone with respect to every argument. That is, if $p_j$ or $h(p_j)$ increases, then so does the overall distance.

If we also ask for a binary function realization of this general function of many arguments, then the binary function must be associative, commutative, differentiable, and monotone. That is, supposing two cases differs at attributes $j_1, \ldots, j_s$, we should have

$$d(\mathbf{c}_i, \mathbf{c}_k) = g(h(p_{i_1}), \ldots, h(p_{i_s}))$$
$$= f(f(\cdots f(h(p_{i_1}), h(p_{i_2})) \cdots), h(p_{i_s})). \tag{3}$$

$h(0)$ should be the identity of $f$. The problem of selecting the right binary function for incremental

combination of individual differences is discussed in Cheng and Kashyap (1989).

## 4 An Entropy Solution

The entropy is a natural connection between $p_j$ and $h(p_j)$. If the cases occur in equal probabilities, then the entropy or the lower bound to the average path length for a decision tree to distinguish them is $\frac{1}{m}\log_2 m$. If an attribute is chosen as the root of the decision tree, and the probability that a case assumes 1 as the value of the attribute is $p$, then the expected average future path length (which is the criterion used in feature selection in many decision tree models) is

$$E(p) = p\left(\tfrac{pm}{m}\log_2 m\right) + (1-p)\left(\tfrac{(1-p)m}{m}\log_2 m\right)$$
$$= (1 - 2p + 2p^2)\log_2 m. \quad (4)$$

This value reaches its minimum $(\log_2 m)/2$ when $p = 1/2$ and maximum $\log_2 m$ when $p = 1$ or $p = 0$.

For two different attributes with probabilities $p_1$ and $p_2$, the expected average future path length is (assuming these attributes are statistically independent)

$$E(p_1, p_2) = (p_1^2 p_2^2 + p_1^2(1-p_1)^2$$
$$+(1-p_1)^2 p_2^2 + (1-p_1)^2(1-p_2)^2)\log_2 m$$
$$= (1 - 2p_1 + 2p_1^2)(1 - 2p_2 + 2p_2^2)\log_2 m \quad (5)$$

In a divisive greedy algorithm of constructing decision trees, the attribute that generates the smallest expected average future path length is considered the strongest candidate, which, in this case, is the one with $p_j$ closest to 0.5. We notice that $E(p)$ is decreasing on $[0,1/2]$, and the combination of it is essentially multiplication. So, a proper $h$ mapping may be

$$h(p) = -\log_2(1 - 2p + 2p^2). \quad (6)$$

Indeed, this $h$ mapping satisfies all the requirements we proposed in the previous section, coupled with the addition as the binary function realization. $h(0) = 0$ is the identity of addition, $h$ is symmetric with respect to 0.5, and the derivative at 0.5 is 0. Thus, $h(p_j) = -\log_2(1 - 2p_j + 2p_j^2)$ can be used as the weight of the $j$th attribute, and it can also be used to measure the dissimilarity between two cases that differ only at the $j$th attribute. In general, the dissimilarity between cases $c_i$ and $c_k$ is

$$d(c_i, c_k) = \sum_{j=1}^{n} h(p_j) e(c_{ij}, c_{kj}). \quad (7)$$

When all $p_j$'s are the same, dissimilarity reduces to the Hamming distance within a multiplicative factor. Replacing $h(p)$ with $-h(p)$, we obtain an additive measure for similarity, which has a logarithmic relation with the recall model $\prod_j e(c_{ij}, c_{kj})^{w_j}$ in the Search of Associative Memory (SAM) model of Raaijmakers and Shiffrin (1981). This is of course not the only formula for dissimilarity. We have to impose an underlying model or other axioms to make this formula unique.

## 5 Demonstration

The Tversky and Gati experiment on country similarity rating now can be modeled using our context-dependent similarity measure. The countries can be represented by certain attributes. We arbitrarily give them names in the tables below, although these attributes obviously are not independent, and some different scenarios may be more appropriate. All numerical computations are based on the formulas presented in the previous section, and the results show that our model solves the similarity rating paradox successfully.

| Country | north | central | communist | neutral |
|---|---|---|---|---|
| Austria | 0 | 1 | 0 | 1 |
| Sweden | 1 | 0 | 0 | 1 |
| Poland | 0 | 0 | 1 | 0 |
| Hungary | 0 | 1 | 1 | 0 |
| $p_j$ | 0.25 | 0.5 | 0.5 | 0.5 |
| $h(p_j)$ | 0.678 | 1 | 1 | 1 |

When we add $h(p_j)$'s of the mismatched attributes, we get the context-dependent dissimilarity between two countries. The following table shows the results.

| $c_i$ | $c_k$ | $d(c_i, c_k)$ |
|---|---|---|
| Austria | Sweden | 1.678 |
| Austria | Poland | 3 |
| Austria | Hungary | 2 |
| Sweden | Poland | 2.678 |
| Sweden | Hungary | 3.678 |
| Poland | Hungary | 1 |

A different context is generated by replacing *Poland* by *Norway*.

| Country | north | central | communist | neutral |
|---------|-------|---------|-----------|---------|
| Austria | 0 | 1 | 0 | 1 |
| Sweden | 1 | 0 | 0 | 1 |
| Norway | 1 | 0 | 0 | 0 |
| Hungary | 0 | 1 | 1 | 0 |
| $p_j$ | 0.5 | 0.5 | 0.25 | 0.5 |
| $h(p_j)$ | 1 | 1 | 0.678 | 1 |

Again, the following are the dissimilarity measures under this context. Notice that now *Austria* is closer to *Hungary* than *Sweden*, the closest country in the previous context.

| $c_i$ | $c_k$ | $d(c_i, c_k)$ |
|-------|-------|---------------|
| Austria | Sweden | 2 |
| Austria | Norway | 3 |
| Austria | Hungary | 1.678 |
| Sweden | Norway | 1 |
| Sweden | Hungary | 3.678 |
| Norway | Hungary | 2.678 |

The Hamming distances (context-independent) between *Austria* and *Sweden* and between *Austria* and *Hungary* are the same. However, different contexts generate different context-dependent similarity ratings, which coincides the finding by Tversky and Gati (1978).

## 6 Changing Context

In many situations, the number of cases is huge, and new cases are added in all the time. The probability $p_j$ must be estimated and updated continuously. This can be done using the following updating rule:

$p_j \leftarrow c_{1j}$
$m \leftarrow 1$
**while** running **do**
  $m \leftarrow m+1; \alpha \leftarrow 1/m$
  $p_j \leftarrow (1-\alpha)p_j + \alpha c_{mj}$

This can be done in parallel for all $j$.

To catch the changing trends in the context, and to avoid keeping track of $m$, the total number of cases, we can stop the decrease of $\alpha$ after certain $m$ is reached. The result is that the ancient statistics will decay gradually. This mechanism can be implemented in the input sensor units in many neural network models to model forgetting in short term or long term memory. These models include the ART clustering and other pattern matching neural networks.

## 7 Concluding Remarks

We propose the axioms and a formula for computing degrees of dissimilarity between objects or cases based on the context, which is defined by the collection of objects or cases. For different contexts, our method produces different similarity structures. This model is used to explain a paradox in similarity rating. Since similarity is a controversial subject in cognitive science and experimental psychology, we need to treat it carefully. Similarity computation is the basis of categorization, clustering, and concept formation. We hope this work will contribute to learning and memory research and benefit knowledge acquisition and analogical reasoning.

In this paper, we simplified and idealized certain situations. Many limitations, for example, the types of attributes, the independence assumption, and the actual form of the formula, can be relaxed through future work.

## References

[1] J.R. Barclay, J.D. Bransford, J.J. Franks, N.S. McCarrell, and K. Nitsch (1974). Comprehension and semantic flexibility. *Journal of Verbal Learning and Verbal Behavior*, **13**, 471-481.

[2] Y. Cheng and R.L. Kashyap (1989). A study of associative evidential reasoning. *IEEE Transactions on Pattern Analysis and Machine Intelligence*, **11**, 623-631.

[3] J.G.W. Raaijmakers and R.M. Shiffrin (1981). Search pf Associative Memory. *Psychological Review*, **88**, 93-134.

[4] A.M. Treisman and G. Gelade (1980). A feature-integration theory of attention. *Cognitive Psychology*, **12**, 97-136.

[5] A. Tversky and I. Gati (1978). Studies of similarity. In E. Rosch and B.B. Lloyd (Eds.), *Cognition and Categorization*. Hillsdale, NJ: Erlbaum.

[6] S. Vosniadou and A. Ortony (1989). Similarity and analogical reasoning: a synthesis. In S. Vosniadou and A. Ortony (Eds.), *Similarity and Analogical Reasoning*, Cambridge: Cambridge University Press.

# Session 2:

# Abductive Probabilistic Reasoning and KB Generation